\documentclass[nohyperref]{article}

\usepackage{microtype}
\usepackage{graphicx}
\usepackage{subfigure}
\usepackage{booktabs} 
\usepackage{makecell}

\usepackage{hyperref}

\usepackage[accepted]{icml2022}

\usepackage{amsmath}
\usepackage{amssymb}
\usepackage{mathtools}
\usepackage{amsthm}

\usepackage[capitalize,noabbrev]{cleveref}

\theoremstyle{plain}

\theoremstyle{definition}

\theoremstyle{remark}

\usepackage[textsize=tiny]{todonotes}

\begin{document}

\twocolumn[
\icmltitle{Pocket2Mol: Efficient Molecular Sampling Based on 3D Protein Pockets}




\begin{icmlauthorlist}
\icmlauthor{Xingang Peng}{thu}
\icmlauthor{Shitong Luo}{helixon}
\icmlauthor{Jiaqi Guan}{uiuc}
\icmlauthor{Qi Xie}{westlake}
\icmlauthor{Jian Peng}{helixon,uiuc,air}
\icmlauthor{Jianzhu Ma}{air,ee}
\end{icmlauthorlist}

\icmlaffiliation{thu}{Tsinghua University, Beijing, China}
\icmlaffiliation{helixon}{HeliXon Limited, Beijing, China}
\icmlaffiliation{uiuc}{Department of Computer Science, University of Illinois at Urbana-Champaign, Champaign, USA}
\icmlaffiliation{westlake}{Westlake University, Hangzhou, China}
\icmlaffiliation{ee}{Department of Electronic Engineering, Tsinghua University, Beijing, China}
\icmlaffiliation{air}{Institute for AI Industry Research, Tsinghua University, Beijing, China}

\icmlcorrespondingauthor{Jianzhu Ma}{majianzhu@tsinghua.edu.cn}

\icmlkeywords{3D pocket, molecule design, auto-regressive, structure}

\vskip 0.3in
]



\printAffiliationsAndNotice{}  

\begin{abstract}

Deep generative models have achieved tremendous success in designing novel drug molecules in recent years. A new thread of works have shown the great potential in advancing the specificity and success rate of {\it in silico} drug design by considering the structure of protein pockets. This setting posts fundamental computational challenges in sampling new chemical compounds that could satisfy multiple geometrical constraints imposed by pockets. Previous sampling algorithms either sample in the graph space or only consider the 3D coordinates of atoms while ignoring other detailed chemical structures such as bond types and functional groups. To address the challenge, we develop Pocket2Mol, an E(3)-equivariant generative network composed of two modules: 1) a new graph neural network capturing both spatial and bonding relationships between atoms of the binding pockets and 2) a new efficient algorithm which samples new drug candidates conditioned on the pocket representations from a tractable distribution without relying on MCMC. Experimental results demonstrate that molecules sampled from Pocket2Mol achieve significantly better binding affinity and other drug properties such as drug-likeness and synthetic accessibility. 


\end{abstract}

\section{Introduction}

Deep learning has achieved great success in drug design \cite{metrics_mol}. One of the representative works is that two out of six compounds designed by Insilico Medicine have been found to be active in cellular assays against kinase DDR1 \cite{insilico}. The main intuition of these generative models is to efficiently represent all the collected chemical structures in a compact low dimensional space and sample new drug candidates by perturbing the hidden values. The outputs of these models vary from 1D chemical descriptors (e.g. SMILES strings \cite{molgen_string}), 2D graphs (e.g. trees \cite{junction_tree, molgen_graph} and hyper-graphs \cite{hypergraph}) to 3D structures \cite{molgen_struc, g_schnet, molgen_struc_2}. 

However, at the molecular level, a small molecule inhibits or activates particular biological functions only by binding to a specific protein pocket. It can be expected that drug design ignoring the target protein pockets could significantly reduce the success rate of downstream cell experiments. Only recently, an increasing amounts of works start to study the pocket-based drug design. To be more specific, given the 3D binding pocket of the target protein, these models are aware of the geometric information of the 3D pocket and generate molecules to bind to the pockets accordingly. 
Early approaches modify the pocket-free models by integrating evaluation functions like docking scores between sampled molecules and pockets to guide the candidate searching \cite{deepligbuilder}. Another types of models transform the 3D pocket structures to molecular SMILES strings or 2D molecular graph \cite{target2smiles, target2smiles_2} without modeling the interactions between the small molecular structures and 3D pockets explicitly. Conditional generative models are developed to model the 3D atomic density distributions within the 3D pocket structures \cite{ligan, AR}. The challenging point of this problem then moves to the efficiency of the structural sampling algorithm from the learned distributions. In addition, previous models over-emphasize the importance of the 3D locations of atoms while ignoring the generation of chemical bonds, which results in unrealistic atomic connections in practice. 

In this work, we improve the pocket-based drug design from the following directions: 1) developing a new deep geometric neural network to accurately model the 3D structures of pockets; 2) designing a new sampling strategy to enable more efficient conditional 3D coordinates sampling; 3) assigning the ability of sampling chemical bonds between a pair of atoms to our model. In particular, we develop an E(3)-equivariant generative model, named Pocket2Mol, leveraging both vector-based neurons and geometric vector perceptrons \cite{vector_neuron, gvp2} as basic building blocks for learning the chemical and geometric constraints imposed by protein pockets. Our model jointly predicts frontier atoms, atomic positions, atom types and chemical bonds with shared atom-level embedding and samples molecules in an auto-regressive way. Thanks to the vector-based neurons, the model can directly generate tractable distributions of the relative atomic coordinates relative to the focal atoms to avoid using the conventional MCMC algorithms \cite{AR}. Experiment results demonstrate that drug candidates sampled by Pocket2Mol not only show higher binding affinities and drug-likeness but also contain more realistic sub-structures than the state-of-the-art models. In addition, Pocket2Mol is much faster than previous auto-regressive sampling algorithms based on MCMC.

To summarize, our key contributions include:
\begin{itemize}
    \vspace{-1.2em}
    \item We construct a new E(3)-equivariant neural network to capture the chemical and geometric constraints imposed by the 3D pockets.
    \vspace{-0.7em}
    \item We propose an efficient conditional molecular sampling algorithm , characterizing a novel position generation strategy and accurate prediction of element types and chemical bonds.
    \vspace{-0.7em}
    \item 
    Drug-likeness is not enough to quantify whether a drug candidate is realistic in biology. We propose a new way to measure whether a unique sub-structure is learned from data or caused by algorithmic bias. 
    \vspace{-0.7em}
    \item
    Our method could sample drug candidates with much better properties than those of the state-of-the-art methods.
\end{itemize}
\section{Related Work}

\paragraph{Molecular Generation based on 3D Protein Pockets}

Here we only review the molecular generation approaches involving protein pockets. \citet{target2smiles} employed a modified GAN model, named BicycleGAN \cite{bicycle_gan}, to represent molecules in the hidden space from protein pockets and decode the representations to SMILES strings using a captioning network. Another representative work by \citet{target2smiles_2} designed two structure descriptors to encode the pocket and generated SMILES using conditional RNN. However, these methods merely generate 1D SMILES strings or 2D molecular graphs. Although they considered the 3D structures of pockets, it is not clear whether the generated compounds based on the form of SMILES strings or graphs could actually fit the geometric landscape of pockets ignoring the 3D structures of compounds.

Another thread of works starts to consider the 3D molecular structures of both pockets and small molecules. \citet{deepligbuilder} proposed a ligand neural network to generate 3D molecular structures and leverage Monte Carlo Tree Search to optimize candidate molecules binding to a specific pocket. Instead of involving 3D pockets in the training process, the function of 3D pockets is to evaluate the quality of drug candidates and guide the generation process. 

Recently, multiple end-to-end learning models are proposed to directly translate 3D pockets to 3D molecular structures with high specificity. \citet{ligan} proposed a conditional variational autoencoder to encode the 3D pockets and generate the voxelized atomic densities for sampling the drug candidates. The model was constructed based on the 3D CNN model which is not equivariant to rotation and hard to scale to large protein pockets. \citet{AR} designed another generative model to estimate the probability density of atom types. The major improvement of this model over previous methods is to choose graph neural networks (GNN) as the backbone and introduce the mask-fill schema in the training process, which enhances the ability to estimate the rough landscape of the conditional probability. However, the model also has its own limitations: 1) it models the atomic densities while ignoring the chemical bonds, which often leads to unrealistic molecular structures; 2) the sampling algorithm relies on MCMC to explore the 3D space, which is sub-optimal and inefficient. 

\paragraph{Vector Feature-based Equivariant Networks}

Another important technique related to our work is the equivariant neural network \cite{cohen2016group,schutt2017quantum,weiler20183d,thomas2018tensor,anderson2019cormorant,fuchs2020se, guan2021energy}. It is common to adopt GNN-based architectures to achieve the global rotation equivariance for 3D objects \cite{klicpera2020directional,en_equivariant}. However, they require the input and hidden features for each layer to be equivariant, which does not fit the vector features such as side-chain angles of each amino acid. To address this problem, 
\citet{vector_neuron} proposes vector neurons which extends 1D scalar neurons to 3D vectors and also defines a set of equivariant operations in the vector space. Later, \citet{gvp2} develops the geometric vector perceptrons to efficiently propagate information between scalar features and vector features. \citet{painn} introduces equivariant message passing for vector representations using equivariant features. 
However, they can only linearly combine 3D vectors to satisfy the rotation equivariance, in essence, limiting the model’s geometric expressiveness.

\paragraph{Generation of Atom Positions} The prediction of the atom positions is the key issue in the auto-regressive 3D molecular generation. A common strategy is to predict the distributions of distances between the new atoms and all previous atoms \cite{molgen_struc, g_schnet} and sample from the joint distributions. However, the long range distances are hard to predict, which introduces additional errors in the atom generation. Another strategy is to build a local spherical coordinate systems and predict the positions in the local space \cite{molgen_struc_2}, but the transformation between Euclidean space and spherical space is inefficient and not straightforward.

\section{Method}

The central idea of Pocket2Mol is to learn a probability distribution of atom or bond type for each location inside the pocket based on the atoms that already exist. To learn this context-specific distribution, we adopt the auto-regression strategy to predict the randomly masked part of the training drug from the rest of it. 

In this section, we first introduce the designed generation procedure in Sec. \ref{sec:generation_procedure} to elicit the modules our model will include. In Sec. \ref{sec:model_arch}, we will describe our graph neural network architecture in details. Specifically, we will show how our model achieve E(3)-equivariance leveraging recent advances and how the encoder module and predictor modules are designed. Finally, in Sec. \ref{sec:training}, we will present the training objectives, by which the model learns which atoms are frontiers, where the new atom should be placed and which element type and bond types should be assigned. 


\subsection{Generation Procedure}\label{sec:generation_procedure}

Formally, the protein pocket is represented as a set of atoms with coordinates $\mathcal{P}^{(\text{pro})}=\{ ({a}^{(\text{pro})}_i, \mathbf{r}^{(\text{pro})}_i) \}_{i=1}^{N}$, where ${a}^{(\text{pro})}_i$ and $\mathbf{r}^{(\text{pro})}_i$ are the $i$th heavy atom identity and its coordinate respectively, and $N$ is the number of atoms of the protein pocket. The molecules are sampled in a continuous way. The already generated molecular fragments with $n$ atoms are denoted as a graph with coordinates $\mathcal{G}^{(mol)}_{n}= \{( {a}^{(\text{mol})}_{i}, \mathbf{r}^{(\text{mol})}_{i} , \mathbf{b}^{(\text{mol})}_{i} )\}_{i=1}^{n}$ where ${a}^{(\text{mol})}_{i}$, $\mathbf{r}^{(\text{mol})}_{i}$ and $\mathbf{b}^{(\text{mol})}_{i}$ represent the $i$th heavy atom, its coordinate and its valence bonds with other atoms, respectively. The model is denoted as $\phi$ and the generation process is defined as follows:
\begin{equation}\label{generation}
\begin{aligned}
    \mathcal{G}^{(\text{mol})}_{n} &= \phi(\mathcal{G}^{(\text{mol})}_{n-1}, \mathcal{P}^{(\text{pro})}), \quad n>1 \\
    \mathcal{G}^{(\text{mol})}_{1} &= \phi(\mathcal{P}^{(\text{pro})}),
\end{aligned}
\end{equation}

\begin{figure}[t]
\begin{center}
\centerline{\includegraphics[width=\columnwidth]{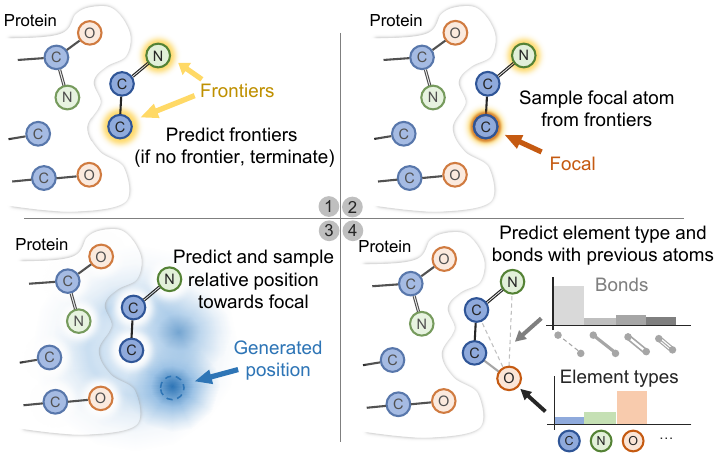}}
\caption{The Generation Procedure of Pocket2Mol. For each panel, the left part is protein and the right part is the sampled molecular fragment.}
\label{fig:generation}
\end{center}
\vspace {-2 em}
\end{figure}

For each atom, the generation procedure consists of four major steps (Fig. \ref{fig:generation}). First, the model's frontier predictor $f_\text{fro}$ will predict the frontier atoms of the current molecular fragments. The frontiers are defined as molecular atoms that can covalently connect to new atoms. If all atoms are not frontiers, it indicates the current molecule is complete and the generation process terminates. Second, the model samples an atom from the frontier set as the focal atom. Third, based on the focal atom, the model's position predictor $f_\text{pos}$ predicts the relative position of the new atom. In the end, the model’s atom element predictor $f_\text{ele}$ and bond type predictor $f_{\text{bond}}$ will predict the probabilities of the element types and the bond types with existing atoms and then sample an element type and valence bonds for the new atom. In this way, the new atom is successfully added into the current molecular fragments and the generation process continues until no frontier atom could be found. Note that this generation process is different for the first atom since there is no molecular atom to be chosen as the frontier yet. For the first atom, all atoms of the protein pocket are used to predict the frontiers and here the frontiers are defined as atoms that the new atoms can be generated within 4 $\rm\AA$. 

\subsection{Model Architecture}\label{sec:model_arch}
Based on the generation procedure mentioned above, the model needs to be composed of four modules: an encoder, a frontier predictor, a position predictor and an element-and-bond predictor. Before we illustrate how each component is designed separately, we will first introduce the E(3)-Equivariant Neural Network (Sec. \ref{sec:building_block}), which we applied as the main architecture to better capture the 3D geometric attributes of pockets and molecular fragments. Following the introduced equivariant network, we will present how the encoder module (Sec. \ref{sec:encoder}) and predictor modules (Sec. \ref{sec:predictors}) are constructed.


\subsubsection{E(3)-Equivariant Neural Network}\label{sec:building_block}
It has been shown that representing the vertices and edges in a 3D graph with both scalar and vector features \cite{en_equivariant, painn} can help boost the expressive power of the neural network. In our network, all vertices and edges in protein pockets $\mathcal{P}^{(\text{pro})}$ and molecular fragments $\mathcal{G}^{(mol)}_{n}$ are associated with both scalar and vector features to better capture the 3D geometric information. In the rest of texts, we use symbols ``$\cdot$'' and ``$\to$'' overhead to explicitly denote scalar features and vector features (e.g. $\dot{\mathbf{x}}$ and $\vec{\mathbf{x}}$). 

We adopt the geometric vector perceptrons \cite{gvp2} and the vector-based neural network \cite{vector_neuron} to achieve E(3)-equivariance. The geometric vector perceptrons (GVP) extend standard dense layers and can propagate information between scalar features and vector features \cite{gvp2}. We briefly formulize the original GVP in the Appendix \ref{ap:sec:gvp} for readers who are not familiar with it. The vector neuron networks expand a set of vanilla neural operations (e.g. linear layer, activation functions) to the vector feature space \cite{vector_neuron}. In particular, we modify the original GVP by adding a vector nonlinear activation to the output vector of the GVP, which is denoted as $G_{\text{per}}$:
\begin{equation}
\begin{aligned}
    (\dot{\mathbf{x}}', \vec{\mathbf{x}}') &= \text{GVP}(\dot{\mathbf{x}}, \vec{\mathbf{x}}) \\
    \vec{\mathbf{x}}' &\leftarrow \text{VectorNonLinear}(\vec{\mathbf{x}}')
\end{aligned}
\end{equation}
where $(\dot{\mathbf{x}}, \vec{\mathbf{x}})$ could be any pair of scalar-vector features. In our model, we choose the LeakyReLU nonlinear functions for both scalar and vector outputs of GVP. In addition, we define a geometric vector linear (GVL) block denoted as $G_{\text{lin}}$ by dropping the nonlinear activations of both scalar and vector outputs of GVP. The modified GVP blocks $G_{\text{per}}$ and the GVL blocks $G_{\text{lin}}$ are the primary building blocks of our model, which enables the model to be E(3)-equivariant.  Besides, including vector features is crucial for our model to directly and accurately predict the atom positions based on the geometric environments provided by the protein pockets (details will be described in Sec. \ref{sec:predictors}).


\subsubsection{Encoder} \label{sec:encoder}

We represent the protein pockets and molecular fragments as a $k$-nearest neighbor (KNN) graph in which vertices are atoms and each atom is connected to its $k$-nearest neighbors. The input scalar features of protein atoms include the element types, the amino acids they belong to and whether they are backbone or side-chain atoms. The input scalar features of molecular atoms include the element types, the valence and the numbers of different chemical bonds. In addition, all atoms have one more scalar feature indicating whether they belong to the protein or the molecular fragment. The scalar edge features include the distances encoded with Gaussian RBF kernels \cite{schnet}, the bond types and a boolean variable indicating whether there is a valence on the edge. The input vector vertex features include the coordinates of atoms, while the vector edge features are the unit directional vector of the edge in the 3D space.

First, multiple embedding layers are applied to embed these features as $(\dot{\mathbf{v}}^{(0)}_i, \vec{\mathbf{v}}^{(0)}_i)$ for vertices and $(\dot{\mathbf{e}}^{(0)}_{ij}, \vec{\mathbf{e}}^{(0)}_{ij})$ for edges. Then $L$ message passing modules ${M_{l}} (l=1,\dots, L)$ and update modules ${U_{l}} (l=1,\dots, L)$ are concatenated interleavedly to learn the local structure representations:
\begin{equation}
\begin{aligned}
(\dot{\mathbf{m}}^{(l)}_i\!, \vec{\mathbf{m}}^{(l)}_i)&=\!\!\!\!\!\!\sum_{j\in \text{KNN}(i)}\!\!\!\! M_{l}(\dot{\mathbf{v}}^{(l-1)}_j\!, \vec{\mathbf{v}}^{(l-1)}_{j}\!, \dot{\mathbf{e}}^{(l-1)}_{ij}\!, \vec{\mathbf{e}}^{(l-1)}_{ij}) \\
(\dot{\mathbf{v}}^{(l)}_i, \vec{\mathbf{v}}^{(l)}_i) &= U_{l}(\dot{\mathbf{v}}^{(l-1)}_i, \vec{\mathbf{v}}^{(l-1)}_i, \dot{\mathbf{m}}^{(l)}_i, \vec{\mathbf{m}}^{(l)}_i)
\end{aligned}
\end{equation}

The message passing module is formulated as:
\begin{equation} \label{eq:message_passing}
\begin{aligned}
    (\dot{\mathbf{v}}_j', \vec{\mathbf{v}}_j') &= G_{\text{lin}}(\dot{\mathbf{v}}^{(l-1)}_j, \vec{\mathbf{v}}^{(l-1)}_j) \\
    (\dot{\mathbf{e}}_{ij}', \vec{\mathbf{e}}_{ij}') &= G_{\text{per}}(\dot{\mathbf{e}}^{(l-1)}_{ij}, \vec{\mathbf{e}}^{(l-1)}_{ij}) \\
    \dot{\mathbf{m}}_j' &= \dot{\mathbf{v}}_j' \circ \dot{F}_{\text{lin}}(\dot{\mathbf{e}}_{ij}') \\
    \vec{\mathbf{m}}_j' &= \dot{F}_{\text{lin}}'(\dot{\mathbf{e}}_{ij}') \circ \vec{\mathbf{v}}_{j}' + \dot{F}_{\text{lin}}''(\dot{\mathbf{v}}_{j}') \circ \vec{F}_{\text{lin}}(\vec{\mathbf{e}}_{ij}') \\
    (\dot{\mathbf{m}}^{(l)}_j, \vec{\mathbf{m}}^{(l)}_j) &= G_{\text{per}}'(\dot{\mathbf{m}}_j', \vec{\mathbf{m}}_j') 
\end{aligned}
\end{equation}
where $\circ$ stands for Hadamard product, $\dot{F}_{\text{lin}}$ and $\vec{F}_{\text{lin}}$ denote the scalar linear layer and the vector neuron linear layer respectively. 
The GVL block $G_{\text{lin}}$ and GVP block $G_{\text{per}}$ work like conventional linear layers and perceptrons to transform both scalar and vector features. The scalar parts of the messages $\dot{\mathbf{m}}_j'$ are the product of scalar features of both vertices and edges. However, the vector messages $\vec{\mathbf{m}}_j'$ cannot be the product of vector features of vertices and edges because this violates the equivariant constraints. Instead, the vector messages are calculated as the summation of vector features of vertices and edges after the multiplication with scalar features, which allows the information interaction between vertices and edges as well as between scalar and vector features. The update module is formulated as:
\begin{equation}
\begin{aligned}
    (\dot{\mathbf{v}}^{(l-1)\prime}_i, \vec{\mathbf{v}}^{(l-1)\prime}_i) &= G_{\text{lin}}'(\dot{\mathbf{v}}^{(l-1)}_i, \vec{\mathbf{v}}^{(l-1)}_i) \\
    \dot{\mathbf{v}}^{(l)}_i &= \dot{\mathbf{v}}^{(l-1)\prime}_i + \dot{\mathbf{m}}^{(l)}_i  \\
    \vec{\mathbf{v}}^{(l)}_i &= \vec{\mathbf{v}}^{(l-1)\prime}_i + \vec{\mathbf{m}}^{(l)}_i
\end{aligned}
\end{equation}
which is a simple addition operation to the messages after going through the GVL block.

\subsubsection{Predictors}\label{sec:predictors}

The encoder functions extract hidden representations $(\dot{\mathbf{v}}_i^{(L)}, \vec{\mathbf{v}}_i^{(L)})$ capturing the chemical and geometric attributes of the protein pockets and molecule fragments. These representations are used for frontier predictor, position predictor and element-and-bond predictor (we omit superscript $L$ in the following for simplicity). 

\paragraph{Frontier Predictor} We define a geometric vector MLP (GV-MLP) as a GVP block followed by a GVL block, denoted as $G_{\text{mlp}}$. The frontier predictor takes the features of atom $i$ as input and utilizes one GV-MLP layer to predict the probability of being a frontier $p_{\text{fro}}$ as follows,
\begin{equation}
\begin{aligned}
    (\dot{p}_{\text{fro}}', \vec{\mathbf{p}}_{\text{fro}}') &= G_{\text{mlp}}^{(\text{fro})}(\dot{\mathbf{v}}_i, \vec{\mathbf{v}}_i) \\
    p_{\text{fro}} &= \sigma(\dot{p}_{\text{fro}}')
\end{aligned}
\end{equation}
where $\sigma$ is the sigmoid activation function.

\paragraph{Position Predictor} The position predictor takes as input the features of focal atom $i$ and predict the relative position of the new atom. Since the vector features are equivariant in our model, they can be directly used to generate the relative coordinates $\Delta\mathbf{r}_i$ towards the focal atom coordinate $\mathbf{r}_i$. We build the output of position predictor as a Gaussian Mixture Model with diagonal covariance $p(\Delta\mathbf{r}_i) = \sum_{k=1}^{K}\pi_i^{(k)}\mathcal{N}(\boldsymbol{\mu}_i^{(k)}, \boldsymbol{\Sigma}_i^{(k)})$ in which parameters are predicted by multiple neural networks as follows:
\begin{equation}
\begin{aligned}
    (\dot{\mathbf{v}}_i^{(\text{pos})}, \vec{\mathbf{v}}_i^{(\text{pos})}) &= G_{\text{mlp}}^{(\text{pos})}(\dot{\mathbf{v}}_i, \vec{\mathbf{v}}_i) \\
    (\dot{\boldsymbol{\mu}}_i', \vec{\boldsymbol{\mu}}_i') &= G_{\text{lin}}^{(\mu)}(\dot{\mathbf{v}}_i^{(\text{pos})}, \vec{\mathbf{v}}_i^{(\text{pos})})  \\
    (\dot{\boldsymbol{\Sigma}}_i', \vec{\boldsymbol{\Sigma}}_i') &= G_{\text{lin}}^{(\Sigma)}(\dot{\mathbf{v}}_i^{(\text{pos})}, \vec{\mathbf{v}}_i^{(\text{pos})})  \\
    (\dot{\boldsymbol{\pi}}_i', \vec{\boldsymbol{\pi}}_i') &= G_{\text{lin}}^{(\pi)}(\dot{\mathbf{v}}_i^{(\text{pos})}, \vec{\mathbf{v}}_i^{(\text{pos})})  \\
    (\boldsymbol{\mu}_i, \boldsymbol{\Sigma}_i, \boldsymbol{\pi}_i) &= \big(\vec{\boldsymbol{\mu}}_i', \exp(\vec{\boldsymbol{\Sigma}}_i'), \sigma_{(\text{sf})}(\dot{\boldsymbol{\pi}}_i')\big)
\end{aligned}
\end{equation}
where $\sigma_{(\text{sf})}$ is the softmax function. After processed by GV-MLP, the mean, covariance and prior probability of Gaussian components are predicted by three separate GVL blocks. Since the vector features are equivariant, the vector outputs of GVL block can directly represent the mean vectors and covariance vectors.

\begin{figure}[t]
\begin{center}
\centerline{\includegraphics[width=\columnwidth]{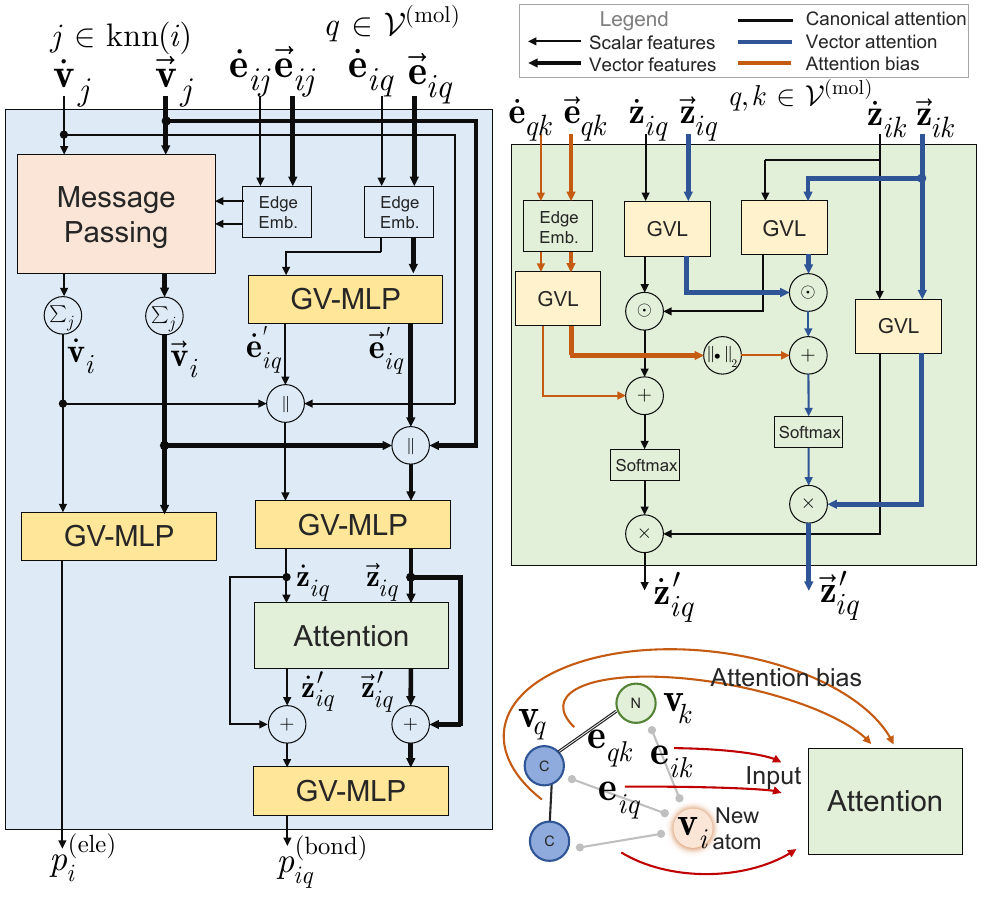}}
\caption{The element-and-bond predictor of the model and the detailed attention module for bond predictions. The symbol $\|$ stands for concatenation and the symbol $\odot$ stands for the inner product operation. The symbol $\|\cdot\|_2$ represents the norm of vectors.}
\label{fig:model}
\end{center}
\vspace {-2 em}
\end{figure}

\paragraph{Element-and-Bond Predictor} After predicting the position of the new atom $i$, the element-and-bond predictor will predict the element type of the new atom $i$ and the valence bonds between atom $i$ and all the atoms $q$ ($\forall q\in \mathcal{V}^{(mol)}$) in the existing molecular fragment. Fig. \ref{fig:model} demonstrates the architecture of the predictor neural network. First, we collect $k$-nearest neighbor atoms $j\in \text{KNN}(i)$ among all atoms. Then a message passing module (Eq. \ref{eq:message_passing}) is utilized to integrate the local information from neighbor atoms into the new atom $i$ position as its representation $(\dot{\mathbf{v}}_i, \vec{\mathbf{v}}_i)$, from which the element type of atom $i$ is predicted. 

In a parallel pathway, the representation of edge between atom $i$ and $q$, denoted as $(\dot{\mathbf{z}}_{iq}, \vec{\mathbf{z}}_{iq})$, is the concatenation of the features of atom $i$, the features of atom $q$ and the processed features of the edge $e_{iq}$, followed by a GV-MLP block, i.e.
\begin{equation}
\begin{aligned}
    \dot{\mathbf{z}}^{(0)}_{iq} &= [\dot{\mathbf{v}}_{i} ; \dot{\mathbf{v}}_{q} ; \dot{\mathbf{e}}'_{iq}]\\
    \vec{\mathbf{z}}^{(0)}_{iq} &= [\vec{\mathbf{v}}_{i} ; \vec{\mathbf{v}}_{q} ; \vec{\mathbf{e}}'_{iq}]\\
    (\dot{\mathbf{z}}_{iq}, \vec{\mathbf{z}}_{iq}) &= G^{(\text{bon})}_{\text{mlp}}(\dot{\mathbf{z}}^{(0)}_{iq}, \vec{\mathbf{z}}^{(0)}_{iq})
\end{aligned}
\end{equation}
where $(\dot{\mathbf{e}}'_{iq}, \vec{\mathbf{e}}'_{iq})$ are the input edge features processed by the edge embedding and one GV-MLP block.

The edge representations are fed into an attention modules to predict the bond types (we regard {\it no bond} as a special bond type). As shown in Fig. \ref{fig:model}, the attention module of Pocket2Mol differs from the conventional attention module in two main aspects: First, scalar and vector features processed by GVL blocks pass through two separate attention pathways. We choose the canonical attention \cite{attention} for scalar features. For the vector features, we propose a new type of attention module defined as follows:
\begin{equation} \label{eq:vector_att}
\begin{aligned}
    a_{qk} &= \langle\vec{\mathbf{z}}_{iq}^{(\text{att})}, \vec{\mathbf{z}}_{ik}^{(\text{att})}\rangle_{F} = \text{tr}\left(\vec{\mathbf{z}}_{iq}^{(\text{att})}\cdot\vec{\mathbf{z}}_{ik}^{(\text{att})\top}\right) \\ 
    a'_{qk} &= \sigma_{(\text{sf})}(a_{qv}) \\
    \vec{\mathbf{z}}_{iq}' &= \sum_{v\in \mathcal{V}^{(\text{mol})}} a'_{qv} \vec{\mathbf{z}}_{iv}^{(\text{att})}
\end{aligned}
\end{equation}
where $\vec{\mathbf{z}}_{iq}^{(\text{att})}, \vec{\mathbf{z}}_{ik}^{(\text{att})}$ and $ \vec{\mathbf{z}}_{iv}^{(\text{att})}$ are the queries, keys and values of vector features, respectively. $\langle\,,\,\rangle_F$ is the Frobenius inner product. The basic rationale of vector-attention module is to generalize the dot product between queries and keys to Frobenius inner product. We prove this operation still keeps the equivariant property in Appendix \ref{app:sec:proof}. Second, we add an attention bias to the product of queries and keys, which is inspired by the triangular self-attention module of AlphaFold \cite{alphafold}, where the attention bias is designed to capture the geometric constraints. The intuition here is shown in the right bottom of Fig. \ref{fig:model}: the attention score of the edge $\mathbf{e}_{iq}$ to the edge $\mathbf{e}_{ik}$ is also related to the edge $\mathbf{e}_{qk}$. For example, if there is a double bond between atom $q$ and $k$, it is nearly impossible that atom $i$ can form valence bond with atom $q$ and $k$ simultaneously. 

\subsection{Training}\label{sec:training}

In the training stage, we randomly mask atoms of molecules and train the model to recover the masked atoms. Specifically, for each pocket-ligand pair, we sample a mask ratio from the uniform distribution $U[0,1]$ and mask corresponding number of molecular atoms. The remaining molecular atoms that have valence bonds to the masked atoms are defined as frontiers. Then the position predictor and the element-and-bond predictor try to recover the masked atoms that have valence bonds to the frontiers by predicting their positions towards corresponding frontiers, the element types and the bonds with remaining molecular atoms. If all molecular atoms are masked, the frontiers are defined as protein atoms that have masked atoms within $4\rm\AA$ and the masked atoms around the frontiers are to be recovered. For the element type prediction, similar with \cite{AR}, we add one more element type representing \textit{Nothing} at the query position. During the training process, we sample not only the positions of masked atoms for element type predictions but also negative positions from the ambient space and assign their labels as \textit{Nothing}.

The loss of the frontier prediction, $\mathcal{L}_{\text{fro}}$, is the binary cross entropy loss of predicted frontiers. The loss of the position predictor, $\mathcal{L}_{\text{pos}}$, is the negative log likelihood of the masked atom positions. For the element type and bond type prediction, we used cross entropy losses for the classification, denoted as $\mathcal{L}_{\text{ele}}$ and $\mathcal{L}_{\text{bond}}$ respectively. The overall loss function is the summation of the above four loss functions:
\begin{equation}
    \mathcal{L} = \mathcal{L}_{\text{fro}} + \mathcal{L}_{\text{pos}} + \mathcal{L}_{\text{ele}} + \mathcal{L}_{\text{bond}}
\end{equation}
We adopt Adam \cite{adam} optimizer to optimize the encoder and all three predictors simultaneously.

\section{Results}

To evaluate the generation performance of Pocket2Mol, we use the CrossDocked dataset \cite{crossdocket} which contains 22.5 million protein-molecule structures and follow the same data preparation and data splitting as \cite{AR} and \cite{ligan}. In the test stage, we randomly sample 100 molecules for each protein pocket in the test set. We compare Pocket2Mol with two baseline methods: a CVAE-based model (CVAE) \cite{ligan} and another auto-regressive generative model (AR) \cite{AR}. \textbf{The hyper-parameters of model architectures and training settings could be found in Appendix \ref{ap:sec:hyper}.} The codes are available at \url{https://github.com/pengxingang/Pocket2Mol}.

\subsection{Evaluation of general properties of sampled molecules}

In this section, we choose metrics that are widely adopted to evaluate the properties of the sampled drug candidates \cite{metrics_mol}: (1) \textbf{Vina Score} estimates the binding affinity between the generated molecules and the protein pockets; (2) \textbf{High Affinity} is calculated as the percentage of pockets whose generated molecules have higher or equal affinities to the ones in the test set; (3) \textbf{QED} is the quantitative estimation of drug-likeness, a value measuring how likely a molecule is a potential candidate for a drug; (4) \textbf{SA} is the synthetic accessibility score, representing the difficulty of drug synthesis (normalized between 0 and 1 and higher values indicate easier synthesis); (5) \textbf{LogP} represents the octanol-water partition coefficient, and in general logP values should be between -0.4 and 5.6 to be good drug candidates \cite{logp}; (6) \textbf{Lipinski} measures how many rules the drug follow the Lipinski's rule of five \cite{lipinski1997experimental, lipinski_variant_Veber2002}, which is an empirical law of drug-likeness; (7) \textbf{Sim.Train} stands for the Tanimoto similarity \cite{tanimoto_sim} with the most similar molecules in the training set; (8) \textbf{Diversity} is calculated as the average Tanimoto similarities of the generated molecules for the individual pockets; (9) \textbf{Time} represents the time cost to generate 100 valid molecules for a pocket. The Vina Score is calculated by QVina \cite{vina} and the chemical properties are calculated by RDKit \cite{rdkit}.

\begin{table}[t]
    \caption{The comparison of general properties of the molecules in the test set and those generated by different methods.}
    \label{tab:gen_metrics}
    \centering

\begin{tabular}{c|cccc}
\toprule
     & \makecell[c]{Test\\Set} & CVAE & AR & \makecell[c]{Pocket2\\Mol} \\ \midrule
    \makecell[c]{Vina Score\\(kcal/mol, ↓)} & \makecell[c]{-7.158\\ $\pm$ 2.10} & \makecell[c]{-6.144\\ $\pm$ 1.57} & \makecell[c]{-6.215\\ $\pm$ 1.54} & {\makecell[c]{\textbf{-7.288} \\ $\pm$ 2.53}} \\ \hline
    \makecell[c]{High Affinity\\(↑)} & - & \makecell[c]{0.238\\ $\pm$ 0.28} & \makecell[c]{0.267\\ $\pm$ 0.31} & {\makecell[c]{\textbf{0.542}\\ $\pm$ 0.32}}  \\ \hline
    QED (↑) & \makecell[c]{0.484\\$\pm$0.21} & \makecell[c]{0.369\\$\pm$0.22} & \makecell[c]{0.502\\$\pm$0.17} & {\makecell[c]{\textbf{0.563}\\$\pm$0.16}} \\ \hline
    SA (↑) & \makecell[c]{0.732\\$\pm$0.14} & \makecell[c]{0.590\\$\pm$0.15} & \makecell[c]{0.675\\$\pm$0.14} & {\makecell[c]{\textbf{0.765}\\$\pm$0.13}} \\ \hline 
    LogP & \makecell[c]{0.947\\$\pm$2.65} & \makecell[c]{-0.140\\$\pm$2.73} & \makecell[c]{0.257\\$\pm$2.01} & {\makecell[c]{\textbf{1.586}\\$\pm$1.82}} \\ \hline
    Lipinski (↑) & \makecell[c]{4.367\\$\pm$1.14} & \makecell[c]{4.027\\$\pm$1.38} & \makecell[c]{4.787\\$\pm$0.50} & {\makecell[c]{\textbf{4.902}\\$\pm$0.42}} \\ \hline
    Sim. Train (↓) & - & \makecell[c]{0.460\\$\pm$0.18} & \makecell[c]{0.409\\$\pm$0.19} & {\makecell[c]{\textbf{0.376}\\$\pm$0.22}} \\ \hline
    Diversity (↑) & - & \makecell[c]{0.654\\$\pm$0.12} & {\makecell[c]{\textbf{0.742}\\$\pm$0.09}} & \makecell[c]{0.688\\$\pm$0.14} \\ \hline
    Time (s, ↓) & - & - & \makecell[c]{19658.56\\$\pm$14704} & {\makecell[c]{\textbf{2503.51}\\$\pm$2207}} \\ \bottomrule
\end{tabular}

\vspace{-1 em}
\end{table}

The mean values of the above metrics are shown in Tab. \ref{tab:gen_metrics}. In general, Pocket2Mol outperforms the other two methods. An interesting observation is that the Vina scores of Pocket2Mol molecules are not only better than other computational models, but also better than the molecules in the test set, indicating that Pocket2Mol has the potential to generate molecules with better affinities with the pockets. In particular, Pocket2Mol successfully generates molecules better than the naturally existing molecules over half of the pockets, almost two-fold higher than the other two approaches. The drug potentials (QED, SA, LogP and Lipinski) are also significantly better than the others, which indicates that molecules generated by Pocket2Mol are more likely to be a drug candiate. In addition, molecules generated by Pocket2Mol have achieved lowest similarities than the molecules in the training set suggesting that Pocket2Mol does not merely memorize the training data. We also observe the diversity of Pocket2Mol is lower than another methods. The diversity metric is commonly adopted to quantify the quality of molecule generation without considering the constraints introduced by the protein pockets. For this pocket-based generation task, a better model capturing the geometric landscape of pockets should not necessarily achieve higher diversity because protein pocket is known to have strong specificity, which is also part of the reason why it often takes decades for biochemists to design one kind of drugs. For instance, for the well-known cancer target protein EGFR (epidermal growth factor receptor), the pocket shape is rough and unique and it is hard to believe that thousands of diverse molecules could fit this pocket \cite{egfr} if considering the chemical and geometrical attributes of the pocket. Therefore, we do not think diversity is an important metric for this task as for many other drug generation tasks. In the end, the generation time of Pocket2Mol is around ten-fold lower than the AR model, which benefits from the direct generation of atom positions instead of randomly exploring the large 3D space using MCMC.

\begin{figure}[t]
\begin{center}
\centerline{\includegraphics[width=\columnwidth]{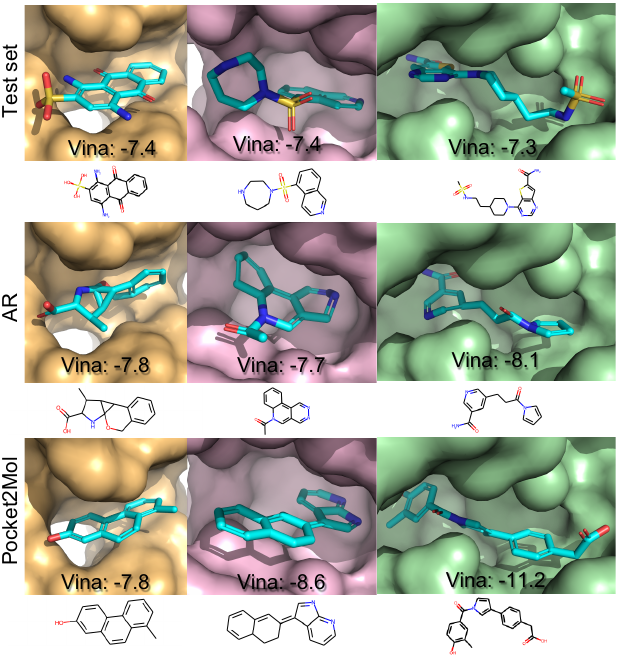}}
\caption{Three examples of generated molecules for pockets. The three columns represent three different pockets (PDB ids are 14GS, 5I0B and 5D7N respectively.)}
\label{fig:cases}
\end{center}
\vspace{-2 em}
\end{figure}

\subsection{Sub-structure analysis}

During evaluation, we find many approaches could achieve reasonable good performance on conventional metrics such as binding affinity and QED but the detailed sub-structures of their sampled molecules are unrealistic.

Therefore, we first visualize the generated molecules and find molecules produced by AR contain numerous distorted benzene rings and excessive three-atom rings (Fig. \ref{fig:cases}). The ring structures composed of three atoms are rare in the dataset (~3$\%$), and both previous approaches generate up to 30$\%$ of the three-atom ring substructures, which is obviously caused by algorithm bias. To systematically investigate this issue, we analyze the ratio of ring-structures with different sizes in the training set and the molecules generated by different approaches. Our hypothesis is that even the chemical structures of the sampled molecules might be different from the test set but the distributions of basic function groups and sub-structures should be maintained, otherwise it would introduce additional difficulty for molecular biologists to understand the biological functions of these molecules.

As shown in Tab. \ref{tab:rings}, the molecules generated by Pocket2Mol show more similar ratios of rings of different sizes with the test set than the other two approaches. In particular, both CVAE and AR are inclined to generate extra amount of three-membered rings, which is caused by adding the chemical bonds after the generation of atoms using OpenBabel \cite{openbabel}. In contrast, Pocket2Mol produces more realistic chemical compounds by jointly learning atom and bond distributions and predicting bonds during the generation process. Another interesting observation is that, for the common five-membered and six-membered rings, Pocket2Mol also achieves most similar ratio to the training and test sets among all three models. 

We further evaluate whether the distributions of the bond angles and dihedral angles of the sampled molecules agree with the test set by using the Kullback-Leibler (KL) divergence. We first find out common bond pairs and bond triples and then use RDKit to calculate the bond angles and the dihedral angles. We also include the KL divergence between the validation set and the test set for reference. As shown in Tab. \ref{tab:bond_angles}, molecules generated by the Pocket2Mol show much lower KL divergence than the other methods, indicating that the molecules generated by Pocket2Mol capture more geometric attributes of data.

\begin{table}[t]
    \caption{The ratio of the molecules containing different rings in the datasets and those generated by different methods.}
    \label{tab:rings}
    \centering
    \begin{tabular}{c|ccccc}
\toprule
    \makecell[c]{Ring\\Size} & \makecell[c]{Train\\Set} & \makecell[c]{Test\\Set} & CVAE & AR & \makecell[c]{Pocket2\\Mol} \\ \midrule
    3 & 0.034 & 0.033 & 0.361 & 0.484 & \textbf{0.002} \\ 
    4 & 0.005 & 0.000 & 0.248 & 0.005 & \textbf{0.000} \\ 
    5 & 0.572 & 0.475 & 0.397 & 0.276 & \textbf{0.415} \\ 
    6 & 0.903 & 0.833 & 0.300 & 0.693 & \textbf{0.885} \\ 
    7 & 0.028 & 0.017 & 0.044 & \textbf{0.033} & 0.076 \\ 
    8 & 0.001 & 0.000 & 0.014 & \textbf{0.007} & \textbf{0.007} \\ 
    9 & 0.000 & 0.000 & 0.006 & 0.006 & \textbf{0.002} \\ \bottomrule
\end{tabular}
    \vspace{-1 em}
\end{table}

\begin{table}[!ht]
    \caption{The KL divergence of the bond angles and dihedral angles with the test set. The lower letters represent the atoms in the aromatic rings.}
    \label{tab:bond_angles}
    \centering
    
\begin{tabular}{c|cc|ccc}
\toprule
     & \makecell[c]{Val.\\Set} & \makecell[c]{Test.\\Set} & CVAE & AR & \makecell[c]{Pocket2\\Mol}  \\ \midrule
    CCC & 0.12  & 0.00  & 7.08  & 1.80  & \textbf{0.97}   \\
    CCO & 0.10  & 0.00  & 7.58  & 2.02  & \textbf{0.95}   \\ 
    CNC & 0.11  & 0.00  & 7.74  & 2.86  & \textbf{0.49}   \\ 
    OPO & 0.10  & 0.00  & 4.72  & 2.06  & \textbf{0.23}   \\ 
    NCC & 0.09  & 0.00  & 7.86  & 2.55  & \textbf{0.95}   \\ 
    CC=O & 0.07  & 0.00  & 7.41  & 2.90  & \textbf{0.76}   \\ 
    COC & 0.12  & 0.00  & 6.32  & 3.88  & \textbf{0.24}   \\ \midrule
    CCCC & 0.14	& 0.00 & \textbf{0.59} & 0.78 & 0.71 \\
    cccc & 0.08	& 0.00 & 7.91 & 10.64 & \textbf{4.49} \\
    CCCO & 0.55 & 0.00 & 0.94 & 1.23 & \textbf{0.56} \\
    OCCO & 1.01 & 0.00 & 1.92 & 1.85 &  \textbf{1.56} \\
    Cccc & 0.28 & 0.00 & 5.78 & 7.91 & \textbf{2.85} \\
    CC=CC & 0.68 & 0.00 & 4.96 & 7.07 & \textbf{4.09}\\
    \bottomrule
\end{tabular}

\end{table}

\begin{figure}[!ht]
\begin{center}
\centerline{\includegraphics[width=\columnwidth]{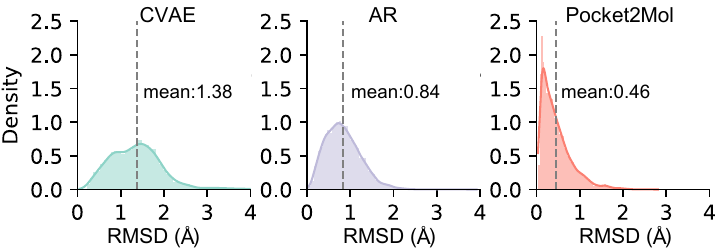}}
\caption{The distributions of RMSD of the generated 3D molecular structures.}
\label{fig:rmsd}
\end{center}
\end{figure}

Moreover, to ensure the validity of generated 3D molecular conformers, we calculate the root-mean-square error (RMSD) between the generated molecular structures and the predicted structures by RDKit. Since a molecule can have multiple potential conformers, we sampled 20 conformers for each molecule using RDKit and choose the one with the minimum RMSD. The distributions of the RMSD of the generated molecules are shown in Fig. \ref{fig:rmsd}. The molecular structures generated by Pocket2Mol generally have lowest RMSD, indicating the correctness of the 3D molecular structures. As discussed in the previous section, accurately predicting the 3D structure of small molecules is crucial for studying their physical and chemistry properties \cite{3d_jiantang}, and more importantly, can determine whether they could actually bind to the 3D pocket. For instance, the algorithm samples a small molecule $A$ with a particular 3D structure designed based on the 3D pocket. However, if $A$ folds to another completely different 3D structure in the living cell, then it is expected the binding affinity of $A$ and the pocket is low.

\section{Conclusion}
Generation of 3D drug-like molecules for protein pockets is a significant but challenging task. In this work, we design Pocket2Mol, an E(3)-equivariant generative network composed of a graph neural network to model the chemical and geometrical features of the 3D protein pockets and a new efficient algorithm to sample new 3D drug candidates conditioned on the pockets. Experiments have demonstrated that molecules generated by Pocket2Mol not only show better affinities and chemical properties but also contain more realistic and accurate structures. We believe that Pocket2Mol will serve as a valuable tool for drug design.


\section*{Software}
All the experiments are conducted on Ubuntu Linux with V100 GPUs. The codes are implemented in Python  3.8 mainly with Pytorch 1.9.0 and our codes are uploaded as Supplementary Material.





\bibliography{references}
\bibliographystyle{icml2022}

\newpage
\appendix
\onecolumn

\section{The formula of the original geometric vector perceptron} \label{ap:sec:gvp}
Here we briefly formulize the original geometric geometric vector perceptron (GVP) and the details can be found in \citet{gvp}. The input and output of GVP are both scalar-vector pairs, denoted as $(\dot{\mathbf{x}}', \vec{\mathbf{x}}') = \text{GVP}(\dot{\mathbf{x}}, \vec{\mathbf{x}})$. The original GVP is defined as:
\begin{equation}
\begin{aligned}
    \vec{\mathbf{x}}_1 &= \mathbf{W}_1 \vec{\mathbf{x}} \,\in\mathbb{R}^{h_1\times 3}\\
    \vec{\mathbf{x}}_2 &= \mathbf{W}_2 \vec{\mathbf{x}_1} \,\in\mathbb{R}^{h_2\times 3}\\
    \dot{\mathbf{s}} &= \|\vec{\mathbf{x}_1}\|_2 \,\in \mathbb{R}^{h_1} \\
    \dot{\mathbf{x}}_1 &= \mathbf{W}_3 \left[\begin{array}{c} \dot{\mathbf{s}} \\ \dot{\mathbf{x}} \end{array}\right] + \mathbf{b}_3 \,\in \mathbb{R}^{h_3} \\
    \dot{\mathbf{x}}' &= \sigma_{\text{act}}(\dot{\mathbf{x}}_1) \\
    \vec{\mathbf{x}}' &= \sigma(\mathbf{W}_g \dot{\mathbf{x}}_1 + \mathbf{b}) \circ \vec{\mathbf{x}}_2\\
\end{aligned}
\end{equation}
where $\mathbf{W}$ and $\mathbf{b}$ are learnable parameters, $ \|\cdot\|$ is row-wise norm and $\circ$ is Hadamard product. $\sigma$ is the sigmoid activation function and $\sigma_{\text{act}}$ is arbitrary activation function which is chosen as Leaky ReLU in our model. We also omit the bias terms in the original GVP.

\section{The hyper-parameters of Pocket2Mol and training}\label{ap:sec:hyper}

The number of message passing layers $L$ are $6$. In the encoder, the hidden dimensions of vertices are fixed as $256$ and $64$ for scalar features and vector features, respectively. The embedding dimensions of edges are $64$ for both scalar and vector features. In the frontier predictor, the hidden dimensions are $128$ and $32$ for scalar features and vector features, respectively. In the position predictors, the hidden dimensions are both $128$ for scalar and vector features because the vector features are important in position prediction. The number of components in the Gaussian Mixture Model is $3$ because we found most atoms in the dataset had no more than three connected atoms. In the element-and-bond predictor, the hidden dimensions are $128$ and $32$ for scalar features and vector features, respectively. The embedding dimensions of edges are $64$ for both scalar and vector features. The numbers of heads of both scalar and vector attentions are $4$. 

We trained Pocket2Mol with batch size $8$ and initial learning rate $2\times 10^{-4}$ and decayed learning rate by a factor of $0.6$ if validation loss did not decrease for $8$ validation iterations. We validated the model every $5000$ training iterations and the number of total training iterations is $475,000$.

\section{The Proof of the equivariant of the vector-attention module}
\label{app:sec:proof}

In the vector-attention module (Eq. \ref{eq:vector_att}), we generalize the vector inner product of queries and keys to Frobenius inner product. We prove that the vector-attention module still keep the equivariance. 

Assume $\vec{\mathbf{z}}_{iq}^{(\text{att})}, \vec{\mathbf{z}}_{ik}^{(\text{att})}$ and $ \vec{\mathbf{z}}_{iv}^{(\text{att})}$ are the original queries, keys and values of vector features and $A$ is the vector-attention module: $\vec{\mathbf{z}}_{iq}' = A(\vec{\mathbf{z}}_{iq}^{(\text{att})}, \vec{\mathbf{z}}_{ik}^{(\text{att})}, \vec{\mathbf{z}}_{iv}^{(\text{att})})$. 
For any rotation or reflection matrix $R\in\mathbb{R}^{3\times 3}$, following Eq. \ref{eq:vector_att} the attention scores are calculated as:
\begin{equation} \label{app:eq:vector_att}
\begin{aligned}
    a_{qk} &= \langle\vec{\mathbf{z}}_{iq}^{(\text{att})}R, \vec{\mathbf{z}}_{ik}^{(\text{att})}R\rangle_{F} \\
    & = \text{tr}\left((\vec{\mathbf{z}}_{iq}^{(\text{att})}R)\cdot (\vec{\mathbf{z}}_{ik}^{(\text{att})}R)^\top \right) \\
    & = \text{tr}\left(\vec{\mathbf{z}}_{iq}^{(\text{att})}RR^\top \vec{\mathbf{z}}_{ik}^{(\text{att})\top} \right) \\
    & = \text{tr}\left(\vec{\mathbf{z}}_{iq}^{(\text{att})} \vec{\mathbf{z}}_{ik}^{(\text{att})\top} \right) \\ 
    & = \langle\vec{\mathbf{z}}_{iq}^{(\text{att})}, \vec{\mathbf{z}}_{ik}^{(\text{att})}\rangle_{F} \\
    a'_{qk} &= \sigma_{(\text{sf})}(a_{qk}) 
\end{aligned}
\end{equation}
which are invariant to the transformation $R$. Then the output vectors of the attention module are:
\begin{equation}
    \vec{\mathbf{z}}_{iq}' = \sum_{v\in \mathcal{V}^{(\text{mol})}} a'_{qv} \vec{\mathbf{z}}_{iv}^{(\text{att})}R = A(\vec{\mathbf{z}}_{iq}^{(\text{att})}, \vec{\mathbf{z}}_{ik}^{(\text{att})}, \vec{\mathbf{z}}_{iv}^{(\text{att})}) R
\end{equation}
which is still equivariant, i.e. $A(\vec{\mathbf{z}}_{iq}^{(\text{att})}R, \vec{\mathbf{z}}_{ik}^{(\text{att})}R, \vec{\mathbf{z}}_{iv}^{(\text{att})}R) = A(\vec{\mathbf{z}}_{iq}^{(\text{att})}, \vec{\mathbf{z}}_{ik}^{(\text{att})}, \vec{\mathbf{z}}_{iv}^{(\text{att})})R$


\end{document}